\documentclass[conference]{IEEEtran}
\IEEEoverridecommandlockouts
\usepackage{cite}
\usepackage{amsmath,amssymb,amsfonts}
\usepackage{algorithmic}
\usepackage{graphicx}
\usepackage{textcomp}
\usepackage{xcolor}
\usepackage{multirow}
\def\BibTeX{{\rm B\kern-.05em{\sc i\kern-.025em b}\kern-.08em
    T\kern-.1667em\lower.7ex\hbox{E}\kern-.125emX}}
\renewcommand{\arraystretch}{1.1}
\begin{document}

\title{Enhancing Molecular Property Prediction via Mixture of Collaborative Experts}

\author{
\IEEEauthorblockN{Xu Yao\textsuperscript{\rm 1}, Shuang Liang\textsuperscript{\rm 1}, Songqiao Han\textsuperscript{\rm 1,2,*}, Hailiang Huang\textsuperscript{\rm 1,2,*}\thanks{\rm * Corresponding author.}}
\IEEEauthorblockA{\textsuperscript{\rm 1}\textit{AI Lab, Shanghai University of Finance and Economics, Shanghai, China}\\
\textsuperscript{\rm 2}\textit{MoE Key Laboratory of Interdisciplinary Research of Computation and Economics}\\
\{yaoxu, liangs1104\}@stu.sufe.edu.cn, \{han.songqiao, hlhuang\}@shufe.edu.cn}
}

\maketitle

\begin{abstract}

Molecular Property Prediction (MPP) task involves predicting biochemical properties based on molecular features, such as molecular graph structures, contributing to the discovery of lead compounds in drug development. To address data scarcity and imbalance in MPP, some studies have adopted Graph Neural Networks (GNN) as an encoder to extract commonalities from molecular graphs. However, these approaches often use a separate predictor for each task, neglecting the shared characteristics among predictors corresponding to different tasks. In response to this limitation, we introduce the GNN-MoCE architecture. It employs the Mixture of Collaborative Experts (MoCE) as predictors, exploiting task commonalities while confronting the homogeneity issue in the expert pool and the decision dominance dilemma within the expert group. To enhance expert diversity for collaboration among all experts, the Expert-Specific Projection method is proposed to assign a unique projection perspective to each expert. To balance decision-making influence for collaboration within the expert group, the Expert-Specific Loss is presented to integrate individual expert loss into the weighted decision loss of the group for more equitable training. Benefiting from the enhancements of MoCE in expert creation, dynamic expert group formation, and experts' collaboration, our model demonstrates superior performance over traditional methods on 24 MPP datasets, especially in tasks with limited data or high imbalance.

\end{abstract}


\section{Introduction}

\begin{figure}
\centering
\includegraphics[width=0.48\textwidth]{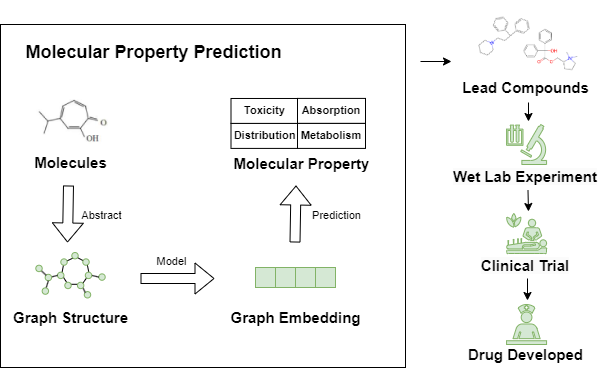} 
\caption{Molecular Property Prediction in Drug Development.}
\label{fig:cycle_of_drug}

\end{figure}

The discovery of lead compounds in drug development often incurs significant time and financial costs \cite{walters2020applications}. Molecular Property Prediction (MPP) plays a crucial role in this process, aiming to identify lead compounds with desired properties within candidate molecules. These properties include factors such as absorption, distribution, metabolism, and excretion (ADME), as emphasized in \cite{huang2021therapeutics}.
Given the potential of deep learning methods in addressing complex problems, numerous approaches based on deep learning have been proposed in recent years for MPP tasks. Many of these methods leverage information, such as molecular structure and elemental composition, to predict potential properties, because the chemical properties of molecules are often closely related to their structure \cite{cahn1966specification}. As depicted in Fig.~\ref{fig:cycle_of_drug}, molecular structures can be naturally represented as graph-structured data, with atoms as nodes and chemical bonds as edges. Hence, studies commonly treat MPP tasks as graph prediction tasks, with our paper specifically emphasizing the graph classification task.

\begin{figure*}[htb]
\vspace{3mm}
\centering
\includegraphics[width=1\textwidth]{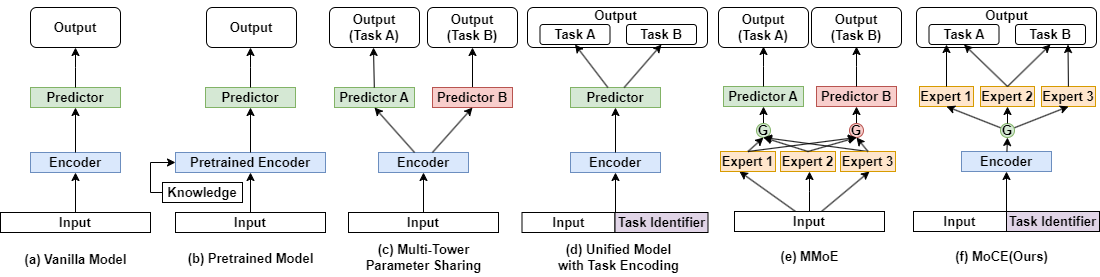} 
\caption{The Model Structures Related to Molecular Property Prediction.}
\label{fig:related_model}
\vspace{1mm}
\end{figure*}

Current graph-based MPP models can be conceptualized as the structure in Fig.~\ref{fig:related_model}(a), comprising two key components: Graph Neural Networks (GNNs) serving as a universal molecular encoder, and standalone feedforward networks acting as a task-specific predictor.
Most of the works explores different GNN structures to enhance the graph representation capability of encoder, such as Gated GNN \cite{li2015gated} or attention-based structures \cite{9412489, yang2019analyzing, guo2020graseq, broberg2022pre, stark20223d}. 
Due to the high cost of biochemical experiments, MPP tasks face issues of data scarcity and imbalance, manifesting as either insufficient overall data or a relative lack of target class data. To overcome these challenges, some works \cite{hu2019strategies} incorporates additional information, like chemical knowledge graphs\cite{fang2022molecular}, into the GNN encoder through pretraining strategies, as illustrated in Fig.~\ref{fig:related_model}(b).  Some other studies discuss the fusion for various tasks, utilizing a multi-tower structure \cite{li2022improving} as Fig.~\ref{fig:related_model}(c), or leveraging inputs with distinct task identifiers \cite{liu2021multi} as  Fig.~\ref{fig:related_model}(d).
However, these methods primarily concentrate on enhancing the common molecular traits by the GNN encoder, often neglecting the exploration of shared elements in predictors corresponding to different tasks.

In this paper, our objective is to enhance the performance of MPP by leveraging the commonalities among molecules and their associations with tasks. We propose the GNN-MoCE architecture based on GNN and Mixture of Collaborative Experts (MoCE), which follows the structure in Fig.~\ref{fig:related_model}(f). In this architecture, GNN serves as an encoder to discover shared traits among molecular structures, while MoCE serves as a universal predictor for various tasks, establishing the connection between molecular prediction and tasks.
The MoCE gives particular emphasis to the homogeneity issue in the expert pool and the decision dominance dilemma within the expert group, which are often overlooked by traditional MoE structures.
On the one hand, MoCE considers the absence of diversity that occurs when experts involved in decision-making receive similar samples and subsequently tend to converge to similar states after training. To handle this problem, we introduce Expert-Specific Projection, leveraging Self-Attention Graph (SAG) pooling \cite{lee2019self} and our Attention Cosine Loss to provide each expert with a unique projection perspective, thereby enhancing expert diversity and allowing experts to unlock collaborative potential within the entire expert pool. 
On the other hand, MoCE also considers the decision dominance within the dynamic decision group. This phenomenon manifests as a specific expert attaining a dominant weight suppresses the learning of other experts in the decision group, leading to the degradation of a decision group into one expert. To alleviate the impact of decision dominance, we propose the Expert-Specific Loss, which emphasizes more equitable training for each expert within the decision group, thereby promoting collaborative effectiveness.

The primary contributions of this study can be summarized as follows:
(1) We present the GNN-MoCE architecture, an efficient approach for molecular graph classification that leverages both GNN and MoCE, and is further enhanced by related tasks. GNN captures common molecular patterns, while our implemented MoCE structure establishes connections between molecular structures and tasks.
(2) During the expert generation, we boost diversity by incorporating the Expert-Specific Projection. The expert decision group is then formed by leveraging task description embedding, which semantically enhances the importance and load loss. To encourage collaboration within the group, we apply our Expert-Specific Loss, ensuring a equitable training for the experts involved in decision-making.
(3) We conducted experiments on 24 datasets, demonstrating the superior performance of our approach, particularly on datasets characterized by limited samples or extreme class imbalance. We have open-sourced the proposed method and all testing datasets, available at: https://github.com/Hyacinth-YX/mixture-of-collaborative-experts.

\section{Related Work}
MPP involves the task of forecasting the physical and chemical properties of molecules based on their internal information. 
Previous research in the field can be categorized into two major classes: the classical method utilizing non-graph features such as quantitative values or sequential information, and the graph-based method employing graph features.

\subsection{Classical Methods in MPP}
Prior to the surge of deep learning, researchers would rely on physical knowledge and employ Density Functional Theory (DFT) calculations, as demonstrated by \cite{orio2009density}, to compute molecular properties. Alternatively, methods such as Quantitative Structure-Activity Relationship (QSAR) \cite{tropsha2010best} utilize theoretical calculations and statistical analysis tools to construct mathematical models that capture the relationship between molecular structure and properties. However, the computational costs associated with DFT and QSAR are both notably high \cite{neese2009prediction}. With the advancement of deep learning, more and more efforts have incorporated deep learning methods to aid in predicting molecular properties \cite{tsubaki2018fast}. 
Some works conducted prediction of molecular properties using sequence modeling, given that molecules could be effectively represented as sequences, such as SMILES expressions \cite{weininger1988smiles} or molecular fingerprints \cite{muegge2016overview}. Recent works, exemplified as \cite{wang2019smiles, li2022deep, pinheiro2020machine}, have demonstrated the utilization of sequence modeling in predicting molecular properties. Specifically, \cite{wang2019smiles} employed the BERT \cite{devlin2018bert} model, while \cite{li2022deep, pinheiro2020machine} utilized CNN \cite{Alex2017CNN} and Feed-Forward Network for sequence modeling, respectively.

\subsection{Graph-based Methods in MPP}
Moleculars can be naturally encoded as graphs, with atoms serving as nodes and bonds forming edges. Since GNN have demonstrated robust representation capabilities for graph-structured data, GNN-based MPP methods have gradually emerged as the mainstream in MPP research. 
As illustrated in Fig.~\ref{fig:related_model}(a), we decompose the model architecture of GNN-based MPP methods into two main components: an encoder and a predictor. The predictor is often a relatively simple linear layer used for result prediction. In contrast, the encoder is the primary focus of existing research, which is commonly viewed as a Message Passing Graph Neural Network (MPNN) \cite{gilmer2017neural} or its variants.
Current research primarily revolves around enhancing the representational capabilities of GNN structures. For instance, \cite{yang2019analyzing} proposed a model called Chemprop based on Directed MPNN (D-MPNN) \cite{dai2016discriminative} to predict molecular properties. Considering the characteristics of MPP tasks, Chemprop incorporates 200 globally calculable molecular features to enhance the performance of MPP tasks.
As the attention mechanism \cite{vaswani2017attention} becomes increasingly prevalent in deep learning, many MPP studies incorporate attention mechanism into GNN models, as seen in works such as \cite{xiong2019pushing, withnall2020building, tang2020self, meng2019property}. In \cite{withnall2020building}, besides attention mechanism, the edge memory schemes are also introduced to further strengthen the representation capability of chemical bonds in molecular graphs. \cite{tang2020self} employs attention mechanism to interpretably learn the relationship between chemical properties and structures. Additionally, \cite{meng2019property} introduces a Gated GNN \cite{li2015gated} structure to enhance graph representation capabilities.

In overcoming the data scarcity and imbalance in MPP, certain endeavors aim to enhance GNN encoder by incorporating additional information. Broadly, these efforts can be categorized into two major classes: pretraining on external information and joint training for diverse internal tasks.

\subsubsection{Enhanced via External Pretraining}

As illustrated in Fig.~\ref{fig:related_model}(b), some research focuses on augmenting molecular embeddings through pretraining graph encoders with novel knowledge. 
Some approaches introduce new views during pretraining. This strategy allows for more effective leveraging of intrinsic molecular characteristics. \cite{jiang2023trangru} employs a combination of Gated Recurrent Unit (GRU) \cite{dey2017gate} and Transformer \cite{vaswani2017attention} structures to capture both local and global information in molecules, facilitating the combination of molecular information at both scales. \cite{guo2020graseq} integrates SMILES and molecular graph representations, achieving an integration of sequence information and molecular graph information through unsupervised reconstruction. \cite{stark20223d} introduces 3D molecular information into the model, endowing the model with the capability to infer 3D molecular structures from 2D information.
Other works related to pretraining focus on incorporating knowledge from diverse domains. For instance, \cite{sun2022pemp} leverages relationships between molecular properties revealed by prior studies in physical chemistry for learning. \cite{broberg2022pre} utilizes knowledge derived from chemical reaction expressions, employing chemical reactions for pretraining.

\subsubsection{Enhanced through Internal Tasks}
Existing research \cite{sun2022does} indicates that current unsupervised pre-training methods offer limited improvement for downstream tasks, falling short of the gains achieved by supervised pre-training.
Supervised learning across various tasks may serve as a supplement to further enhance the effectiveness of MPP. 
A frequently employed architecture is the multi-tower model with parameter sharing, depicted in Fig.~\ref{fig:related_model}(c). This structure is utilized by \cite{li2022improving} in their work. The shared encoder processes input molecules and encodes them into a unified representation across various tasks. The model concurrently performs multiple tasks, providing diverse outputs, from which users can select the relevant portions.
Another variant structure is illustrated in Fig.~\ref{fig:related_model}(d), as employed by \cite{liu2021multi} in their work with the task encoding. This approach employs one-hot encoding to map different tasks into a vector space, acquiring task embeddings through a GNN. By incorporating a task identifier in the input, the model can generate a unified output representing results for different tasks, thereby gaining the capability to leverage inter-task relationships.

In summary, the existing research has primarily focused on enhancing the expressive capabilities of GNN for molecular representation, aiming to improve the effectiveness of MPP tasks. However, there is still an insufficient exploration into how models from diverse tasks can synergize to enhance ensemble effects. Based on MoE, our approach can be viewed as a unified model for integrating sub-models trained on diverse tasks, providing a basis for discussing the collaboration among sub-models.

\subsection{MoE Structure}
MoE \cite{shazeer2017outrageously} has demonstrated distinct advantages in leveraging task distinctions to extract specific task nuances. An example is the Multi-gate Mixture-of-Experts (MMoE) \cite{ma2018mmoe/10.1145/3219819.3220007}, which is depicted in Fig.~\ref{fig:related_model}(e). In contrast to the classical parameter sharing models, MMoE employs multiple gate layers to route features to different tasks. As a result, MMoE can effectively discriminate between different tasks.
Unlike the use of MoE in the encoder as MMoE, we retain the GNN structure in our encoder, and replace the task-specific predictors with a shared expert pool. This allows us to facilitate task collaboration using simple Multi-Layer Perceptron (MLP) experts.

\begin{figure*}[bt]
    \centering
    \includegraphics[width=0.9\linewidth]{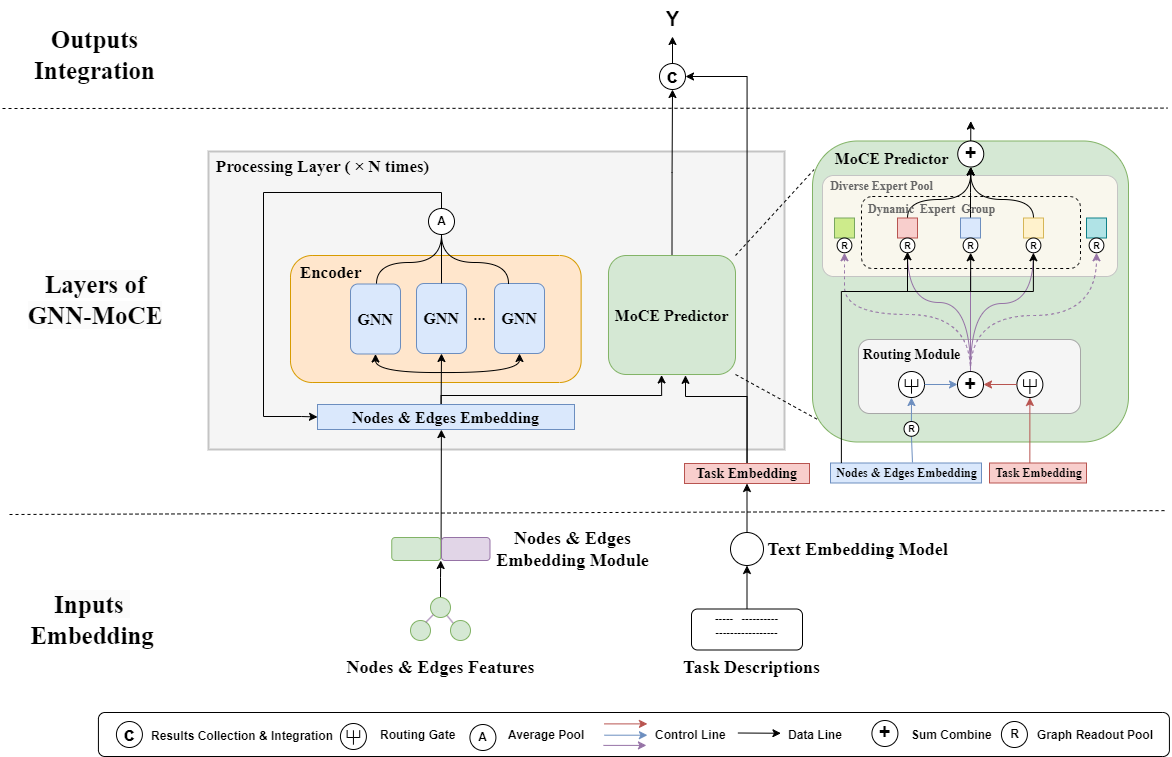}
    \caption{Architecture of GNN-MoCE}
    \label{fig:overall-structure}
\end{figure*}

\section{Architecture}

\subsection{Problem Statement}
The objective of MPP is to predict molecular properties, such as toxicity or activity. A dataset with a size of $N$ can be represented as $\{(x_i, y_i, t_i)\}_{i=1}^N$, comprising molecular graph features $x_i$, corresponding labels $y_i$, and task description embeddings $t_i$. Molecular graph features $x_i$ encompass node and edge attributes, such as atomic types and bond types. 

\subsection{Network Architecture}
Expanding on the traditional message-passing GNN structure, we introduce an architecture named GNN-MoCE. Diverging from other studies that primarily focus on improving graph embeddings, our emphasis lies in the collaboration within the MoCE.
Here, GNNs are solely responsible for embedding graph information, while the experts in the MoCE layer are assigned the task of predicting task-specific outcomes. 
Fig.~\ref{fig:overall-structure} illustrates the overall architecture of GNN-MoCE. 

\subsubsection{Inputs Embedding}
The model inputs consist of two components: molecular graph information and task information. The molecular graph information encompasses common features of atoms and chemical bonds, while the task information is derived from the natural language descriptions of tasks in the dataset. By embedding the natural language descriptions of tasks, we can obtain the semantic representation of tasks, which plays a crucial role in forming the dynamic expert decision group. This is because experts who perform similar tasks are more likely to achieve better performance through collaboration, and task semantics provide a good perspective for measuring the degree of task similarity.
These features are fed into the stacked processing layers of GNN-MoCE. Each layer undertakes the tasks of graph embedding and result prediction, which correspond to the GNN Encoder and MoCE Predictor within the component, respectively.

\subsubsection{GNN Encoder}
GNN Encoder consists of multiple GNN modules based on the message-passing structure. In our work, we configure the GNN module as Graph Isomorphism Network (GIN) layers \cite{xu2018powerful}. During the message propagation phase, the state of graph nodes is iteratively updated based on the states of their neighboring nodes and the features of connecting edges. For a given node \(v\), its hidden state \(p^{(t+1)}\) at time \(t+1\) is defined as follows:
\begin{equation} \label{eq:message_pass1}
    m^{(t+1)}_v = \sum_{w \in N(v)} M_t(p^t_v, p^t_w, e_{vw}),
\end{equation}
\begin{equation} \label{eq:message_pass2}
    p^{(t+1)}_v = U(p^t_v, m^{(t+1)}_v).
\end{equation}
Where \(N(v)\) represents all neighboring nodes of \(v\), and \(m_v^{(t+1)}\) is the \(t+1\) layer message that aggregates embeddings of all neighboring nodes and edge embeddings. 
$p_i^t$ and $e_{ij}$ represent the embeddings of node $i$ and edge $ij$, respectively.
Using the update function \(U\), we obtain embeddings for the \(t+1\) layer. 
Theoretically, the GNN Module can be replaced with other pre-trained GNN embedding modules proposed in the existing literature.

\subsubsection{MoCE Predictor}
The MoCE Predictor, designed based on the MoE structure, makes decisions for samples by the dynamic decision group. Each prediction involves assigning appropriate experts from the expert pool to form the expert group. We will introduce the predictor in the following three parts: the input of the expert group, the formation of the expert group, and the weighted voting of the expert group.

\textbf{The inputs of the expert group} include the node embeddings of the graph and the task description embeddings. The node-level embeddings of the $i$-th graph need to be read out as graph-level information $\hat x$ through a graph readout module $R$ first.
\begin{equation}
    \hat x_i = R(x_i).
\end{equation}

In the MoCE architecture, we employ two distinct graph readout modules. The first one operates before the sample is routed to an expert, requiring the embedding of nodes through Global Mean Pooling (GMP) to obtain a graph-level embedding. This enables the routing module to allocate the sample to an appropriate expert based on the overall characteristics of the graph. The second readout module, referred to as SAG pooling, is associated with each expert. In this context, some unique parameters are employed for each expert, thereby guaranteeing that each expert possesses an individualized graph-reading mechanism. Details of the latter are elaborated in Section \ref{part:diversity}.

\textbf{The formation of the expert group}\label{part:group_forming} can also be viewed as the process of routing the sample to the top-$k_s$ experts. This process is related to both the graph information and task information of the sample.
The $i$-th sample consists of two parts: the graph-level readout embedding $x_i \in \mathbb{R}^{e_f}$ and the task description embedding $t_i \in \mathbb{R}^{e_t}$. Here, $e_f$ and $e_t$ represent the dimensions of graph features and task embeddings, respectively. Upon receiving the current layer's graph and task embedding, the MoCE's routing module directs samples to the appropriate experts.
Gating scores $H(\hat x_i, t_i) \in \mathbb R ^ m$ for pre-defined $m$ experts are computed using graph-level embedding $\hat x_i$ and task embedding $t_i$. We denote $h_{ij}$ as the score of sample $s_i$ for the $j$-th expert. Following the noise mechanism described in \cite{shazeer2017outrageously}, we assume $h_{ij} \sim \mathcal{N}(\mu_{ij}, \sigma_{ij})$, where $\mu_{ij}$ and $\sigma_{ij}$ are estimated as follows:
\begin{equation}
    \mu_{ij} = ( \Gamma ( W_{\mu_1}^T \hat x_i ) + W_{\mu_2}^T t_i )_j ,
\end{equation}
\begin{equation}
    \sigma_{ij} = \text{Softplus}( \Gamma ( W_{\sigma_1}^T \hat x_i ) + W_{\sigma_2}^T t_i )_j .
\end{equation}
Where $W_{\mu_1}, W_{\sigma_1} \in \mathbb{R}^{e_f \times m}$ are learnable parameters for mapping sample feature embeddings, and $W_{\mu_2}, W_{\sigma_2} \in \mathbb{R}^{e_t \times m}$ are learnable parameters for mapping task embeddings. $\Gamma(\cdot)$ is a specific mapping function that retains the top $k_t$ values in a vector and fills the rest with the minimum value. This function is used to constrain expert selection to be within the top $k_t$ experts preferred by the task, while still allowing some flexibility for routing to experts preferred by other samples.

Therefore, $h_{ij}$ can be expressed as follows, where $StdNormal()$ represents random value generated from a standard normal distribution.
\begin{equation}
    h_{ij} = \mu_{ij} + \sigma_{ij} \cdot \text{StdNormal}().
\end{equation}
After obtaining $h_{i*} = [h_{i1}, h_{i2}, \dots, h_{im}]$, we retain the top $k_s$ values and set the rest to zero, resulting in the final gate $G(x_i, t_i)$ of sample $i$:
\begin{equation} \label{eq:gi_softmax}
    G(x_i, t_i) = \text{Softmax}( \text{KeepTopK}(h_{i*}) ).
\end{equation}

\begin{figure}
    \centering
    \includegraphics[width=1\linewidth]{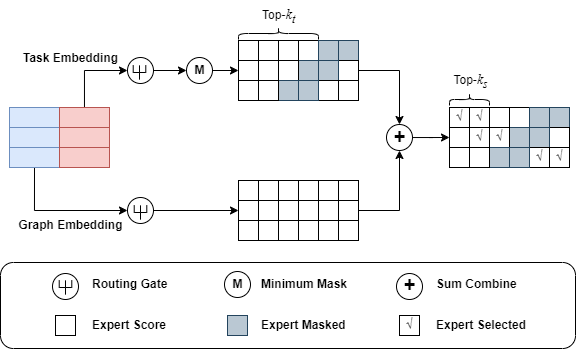}
    \caption{The process of routing samples with task information to experts.}
    \label{fig:task_routing_machenism}
\end{figure}

The aforementioned routing mechanism is visually depicted in Fig.~\ref{fig:task_routing_machenism}. Both the graph embedding and task embedding undergo a routing gate to be mapped into expert scores. For task routing scores, with the existence of the minimum mask, experts outside of the top-$k_t$ are filled with the minimum expert score. After adding the task routing scores to the sample routing scores, the final routing score matrix is inclined towards the top-$k_t$ experts. This achieves the adjustment of expert selection based on the task embedding. Based on the final routing score matrix, the top-$k_s$ experts with high routing scores will be assigned to this sample.

\textbf{The weighted voting of the expert group} is implemented based on the gates calculated above, and the expert's corresponding gate is essentially its weight in the voting process. Let $g_j(\hat x_i, t_i) \in G(\hat x_i, t_i)$ represent the gate scalar for sample $i$, and $E_j(\tilde x)$ denote the output of the $j$-th expert for this sample. We can obtain the 
 integrated result from $m$ experts in this MoCE layer as:
\begin{equation} \label{eq:layer_output}
    y_i = \sum_{j}^{m} g_j(\hat x_i,t_i) E_j(\tilde x_{ij}).
\end{equation}
It is noteworthy that each expert is assigned distinct learnable parameters of SAG readout module. Hence, the graph input for expert $j$ in \eqref{eq:layer_output} is actually the graph representation $\tilde x_{ij} = R_j(x_i)$ obtained after applying its own readout. 

\subsubsection{Outputs Integration}
Each component predicts the results based on the molecular embeddings obtained at that layer, and these predictions need to be collected and integrated at the end. As model's preferences for information at each layer may vary across tasks, the aggregation of these results needs to be adjusted based on the task embedding information. 

For the \(i\)-th sample,  its corresponding task embedding is denoted as \(t_i (i \in 1,\dots,n)\). Let $Map(*)$ represent the mapping function that assigns weights to results based on the task embedding. We define $W_i$ as the mapping weight of $i$-th sample.
\begin{equation}
W_i = \text{Softmax}(\text{Map}(t_i)).
\end{equation}
Consequently, the final output is given by,
\begin{equation}
r_i = \sum_{l}^m ( w_{il} o_{il} ).
\end{equation}
In this formula, \(w_{il} \in W_i\) corresponds to the weight of the \(l\)-th layer. \(o_{il}\) denotes the result of the $l$-th MoCE layer for the \(i\)-th sample, where \(l\) ranges from 1 to \(L\), representing the total layers number.

\section{Collaborative Experts}\label{part:component_ftmoe}

\begin{figure}
    \centering
    \includegraphics[width=0.99\linewidth]{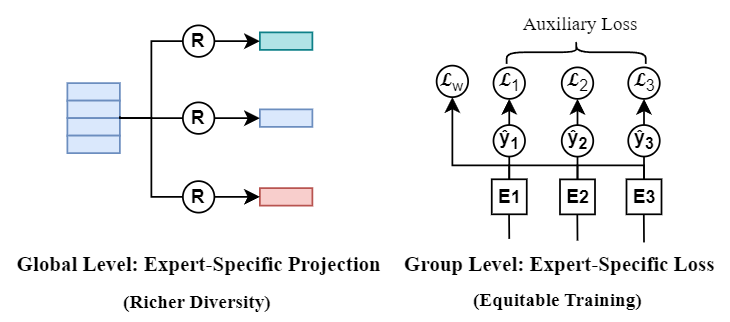}
    \caption{Methods for Collaborative Experts.}
    \label{fig:collaboration}
\end{figure}

The traditional Sparsely Gated MoE structure lacks consideration for collaboration among experts. This deficiency can be viewed from both global and local perspectives. Globally, experts share the same input, and the distinctions between experts primarily rely on the random initialization of parameters. This results in experts having very similar parameters, leading to unstable performance and difficulty in achieving ensemble benefits. Locally, we define the dynamic combination of the top-$k_s$ experts that collectively vote on a particular sample as a dynamic expert group. We observe that traditional MoE structures often exhibit a dominant expert with extremely high weights in the expert group. This dominance suppresses the learning of other experts in the group, making the decision group degrade to one single expert and diminishing the effectiveness of collaboration. To address these issues and enhance expert collaboration on both global and expert group levels, we propose two methods: Expert-Specific Projection and Expert-Specific Loss. Fig.~\ref{fig:collaboration} depicts these two methods.

\subsection{Global Collaboration} \label{part:diversity}
\begin{figure}
    \centering
    \includegraphics[width=1\linewidth]{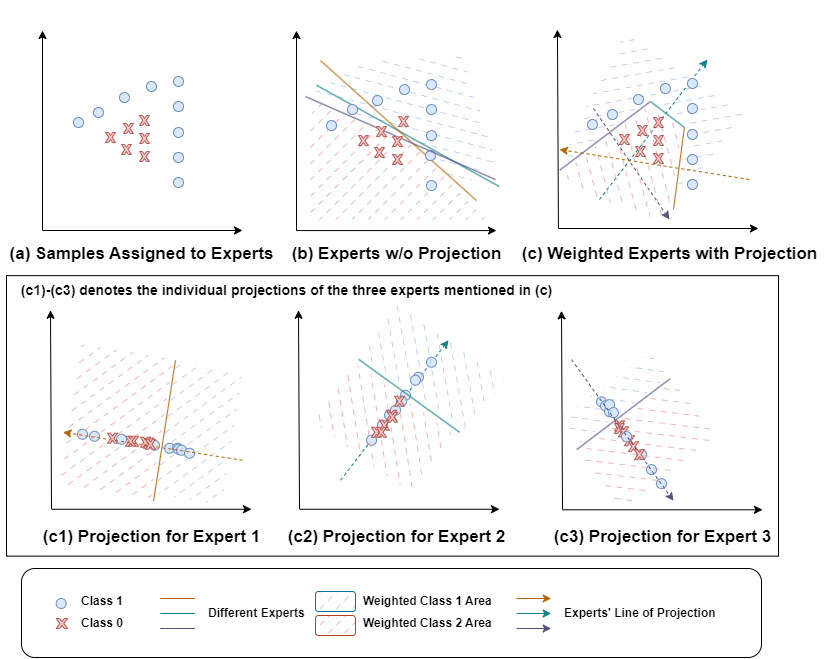}
    \caption{Projection of Experts.}
    \label{fig:expert_diversity_importance}
\end{figure}

The classical MoE structure exhibits a limitation in providing diverse inputs to individual experts, potentially leading to limited diversity among experts. This consequently impacts the global collaboration effectiveness of experts. To address this issue, we propose the method of Expert-Specific Projection, assigning each expert its unique perspective on viewing samples. In Fig.~\ref{fig:expert_diversity_importance}, we intuitively illustrate the significance of expert diversity and explain how our Expert-Specific Projection contributes to enrich expert diversity. 
Fig.~\ref{fig:expert_diversity_importance}(a) depicted a simple binary classification scenario. In a conventional MoE structure, the group of  experts make decisions together may be allocated to similar sample spaces, leading to a learned state illustrated in Fig.~\ref{fig:expert_diversity_importance}(b). It is evident that the learning of experts is constrained by the diversity of samples, making it challenging to achieve improved ensemble performance through voting.
Contrastingly, by assigning distinct projection planes to each expert, enabling them to learn classifiers on these planes, we obtain diverse classification planes that each expert may fit, as illustrated in Fig.~\ref{fig:expert_diversity_importance}(c1-c3). The integration of these experts is depicted in Fig.~\ref{fig:expert_diversity_importance}(c), demonstrating superior weighted classification performance compared to traditional MoE. Therefore, by making the projection planes distinct, we can enrich expert diversity as much as possible, fostering collaboration among experts. Our Expert-Specific Projection method is designed following this rationale.

In detail, we utilize the SAG pooling approach \cite{lee2019self} to project the generic node representations obtained from the graph encoder onto individual perspectives of each expert, thereby obtaining distinctive graph-level representations. In contrast to traditional global graph readout methods like Global Mean Pooling (GMP) \cite{lin2013network}, SAG pooling has the capability to selectively read graph information based on a learnable vector $\Theta_{att} \in \mathbb{R}^{F \times 1}$. Specifically, the self-attention scores $\tilde Z \in \mathbb{R}^{N \times 1}$ for each node on the graph are computed as follows:
\begin{equation} \label{eq:sag_score}
    Z = X \Theta_{att}
\end{equation}
\begin{equation} \label{eq:sag_atilde}
    \tilde A = A + I_N
\end{equation}
\begin{equation} \label{eq:sag_fit}
    \tilde Z = \sigma(\tilde{D}^{-\frac{1}{2}} \tilde A \tilde D^{-\frac{1}{2}} Z)
\end{equation}

In \eqref{eq:sag_score}, $X \in \mathbb{R}^{N \times F}$ represents the feature embedding matrix for nodes, where $N$ is the number of nodes and $F$ is the feature dimension. Equation \eqref{eq:sag_score} is the main component for calculating attention scores, resulting in $Z \in \mathbb{R}^{N \times 1}$ as the basic attention scores for each node under the given $\Theta_{att}$. However, due to the graph's unique structure, we need to adjust the attention scores to incorporate more topological information using the adjacency matrix $A \in \mathbb{R}^{N \times N}$. The adjacency matrix with self-connections $\tilde A$ is obtained as \eqref{eq:sag_atilde}. Additionally, $\tilde D$ represents the degree matrix corresponding to $\tilde A$, and $\sigma$ denotes the activation function. Equation \eqref{eq:sag_fit} produces the attention scores adjusted for the graph structure.

Subsequently, a pooling ratio $\kappa \in (0, 1]$ is defined to adjust the number of subgraph nodes to be selected. Based on the obtained attention scores $\tilde Z$, the top $\lceil \kappa N \rceil$ nodes are selected, and their feature embeddings are aggregated, e.g., by computing their mean. The aggregated result serves as the representation of the graph under a given $\Theta_{att}$.

The reason for employing the SAG pooling method is that it enables the control of the expert’s perspective in observing the samples by manipulating the value of $\Theta_{att}$, thereby enhancing the capture of local details in the graph. On the one hand, for information comprehensiveness, they can collectively obtain information about the entire graph when there are a sufficient number of experts. On the other hand, for information diversity, differentiating the $\Theta_{att}$ values can make the inputs to the experts more diverse. A variety of expert input views can further improve the overall performance of the system. 

To ensure that the perspectives of the experts are as distinct as possible, we add an Attention Cosine Loss term $\mathcal L_{att}$ to the overall loss. For M experts and an embedding dimension of F, we normalize $\Theta_{att_i}$ as follows:
\begin{equation}
    \tilde \Theta_{att_i} = \frac{\Theta_{att_i}}{||\Theta_{att_i}||_2}
\end{equation}
Then, we define the Attention Cosine Loss term as:
\begin{equation}
    \mathcal L_{att} = \frac{1}{M^2} \sum_i^M \sum_j^M \tilde \Theta_{att_i}^T \tilde \Theta_{att_j}.
\end{equation}
Here, $\frac{1}{M^2}$ is a scaling factor to stabilize the numerical values of the loss term. 

\subsection{Group Collaboration}

\begin{figure}
    \centering
    \includegraphics[width=1\linewidth]{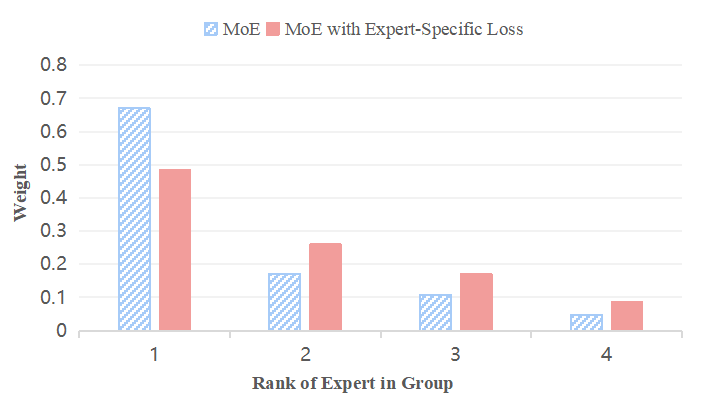}
    \vspace{-7mm}
    \caption{Voting Weights of Experts in Decision Group}
    \label{fig:expert_group_weight}
\end{figure}

In our research, we observed the presence of a dominant expert with extremely high weights in the expert group of the classical MoE structure. This dominant expert hinders the training of other experts within the same group, thereby affecting the overall decision-making effectiveness of the expert group. To address this issue, we introduce the Expert-Specific Loss, ensuring that each expert within the group undergoes fair training. In Fig.~\ref{fig:expert_group_weight}, we illustrate the voting weights of experts in the decision group in traditional MoE and MoE with our proposed loss, respectively. It is evident that in the original MoE structure, one expert typically occupies close to 80\% of the weight, with only 20\% of the weight distributed among other experts. However, upon incorporating our Expert-Specific Loss, each expert receives equitable training, leading to an increase in the weights of other experts. This results in a more significant contribution to the decision-making process and prevents the degradation of the expert group into a single expert.

To be more specific, rather than solely optimizing the final weighted output like other works \cite{6797059, shazeer2017outrageously, du2022glam}, we integrate the individual loss of each expert into the loss for ensemble weighted outputs. In other words, this entails enhancing the individual task performance of each expert.
Formally, we augment the ultimate optimization loss by introducing the Expert-Specific Loss $\mathcal{L}_{exp}$. This term is defined as follows:
\begin{equation}
    \mathcal L_{exp} = \sum_i^M \sum_j^N \mathbb I(i,j) BCE(\hat y_{ij}, y_j).
\end{equation}
Where $N$ and $M$ represent the number of records and the number of experts, respectively, and the indicator function $\mathbb I(i,j)$ indicates which data point is assigned to which expert.

\subsection{Overall Losses}

The enhancement of expert collaboration in our model is primarily achieved by incorporating auxiliary losses. In summary, our Collaborative Loss comprises the following components:
\begin{equation}
    \mathcal L_{col} = \mathcal L_{att} + \mathcal L_{exp} + \mathcal L_{imp} + \mathcal L_{lod}.
\end{equation}
Subsequently, the Collaborative Loss is multiplied by a scalar $\beta \in (0,1]$ and then added to the model's original base loss $\mathcal L_{base}$. This operation yields our ultimate loss term, denoted as $\mathcal L_{overall}$:
\begin{equation}
    \mathcal L_{overall} = \mathcal L_{base} + \beta \cdot \mathcal L_{col}.
\end{equation}
In addition to the aforementioned $\mathcal L_{att}$ and $\mathcal L_{exp}$, we introduce two supplementary loss $\mathcal L_{imp}$ and $\mathcal L_{lod}$ to balance the samples and tasks distributed to experts. 
The introduction of these two additional auxiliary losses is motivated by the imbalance in the global allocation of samples and tasks, where a minority of experts may be assigned a disproportionate amount of samples. This phenomenon has been described in \cite{Eigen2013LearningFR} and \cite{Bengio2015ConditionalCI}. To address this imbalance issue, \cite{shazeer2017outrageously} proposed two types of losses, namely Importance and Load losses. Building upon this method, we incorporate task information into these two losses, ensuring a more balanced allocation of both tasks and samples.
Although they are not the core components of our study, they play a complementary role in improving the model's effectiveness.

\subsubsection{Task-adjusted Importance Loss}
With the computed gate $G(x, t)$ in section \ref{part:group_forming}, we determine the importance of each expert as the sum of its gate's score in a batch. Then, we balance the distribution of expert selection by minimizing the coefficient of variation of expert importance, which is specifically defined as the Importance loss $\mathcal L_{imp}$:

\begin{equation}
    I_j = \sum_{i}G(x_i, t_i)_j
\end{equation}
\begin{equation}
\mathcal L_{\text{imp}} = \beta \cdot \text{CV}([I_1, I_2,\dots,I_m])^2
\end{equation}

Where $\beta$ serves as a scaling factor for the loss, and CV represents the coefficient of variation. This loss term ensures a more balanced distribution of expert weights within the same batch.

\subsubsection{Task-adjusted Load Loss}
The load balance loss term $\mathcal{L}_{lod}$ is used to balance the number of samples each expert receives. This is achieved by balancing the probabilities of experts selecting samples. Since a sample $i$ is chosen by expert $j$ only when $h_{ij}$ is the $k$-th greatest element in $h_{i\cdot}$, it implies that when $h_{ij}$ exceeds a threshold $h_{k_{ij}}$ (representing the k-th largest value in $h_{i\cdot}$ but excluding $h_{ij}$), the probability of sample $i$ being chosen by expert $j$ can be defined as:
\begin{equation}
    P(x_i, t_i)_j = Pr(h_{ij} > h_{k_{ij}}).
\end{equation}
As $h_{ij}$ is modeled in our system to follow $\mathcal{N}(\mu_{ij}, \sigma_{ij})$, this probability can be expressed as:
\begin{equation}
    P(x_i, t_i)_j = \Phi(\frac{\mu_{ij} - h_{k_{ij}}}{\sigma_{ij}}).
\end{equation}
Similar to the definition of the Importance loss, we use the coefficient of variation to define the final Load loss:
\begin{equation}
    L_j = \sum_i P(x_i, t_i)_j
\end{equation}
\begin{equation}
    \mathcal L_{lod} = \beta \cdot CV([L_1, L_2, \dots, L_m])
\end{equation}

In summary, through the optimization of importance and load loss with task parameters, we facilitate sample routing based on task information.
This approach enables a broader participation of experts in tasks, thus reducing the risk of these experts degenerating into a tiny subset.

\section{Experiments}

\subsection{Experimental Setting}
\subsubsection{Datasets}
We conducted experiments on 24 MPP datasets sourced from Therapeutics Data Commons \cite{huang2021therapeutics}. These encompass diverse tasks such as ADME (absorption, distribution, metabolism, and excretion) properties, Tox Prediction, and High-throughput Screening Prediction. 
Given that Tox21 is a multi-label dataset, we partitioned the dataset into 12 subtasks, resulting in training and validation across 35 tasks. In pursuit of conciseness, for the Tox21 dataset, we only report the average effectiveness across different tasks of Tox21.
We conducted separate training for baseline methods, while jointly trained on all 35 tasks for our GNN-MoCE model.

For each dataset, we all applied stratified scaffold split \cite{yang2019analyzing} to get the develop (80\%) and test (80\%) datasets. Within the develop Datasets, the train and validation datasets respectively constitute 80\% and 20\% of the total. The scaffold split is a method of dataset partitioning based on molecular scaffolds. In comparison to a purely random split, tasks generated through scaffold split are more challenging, demanding higher model generalization capabilities, and providing a better approximation to real-world data scenarios. The stratified scaffold split we employ takes into account datasets with limited quantities or significant imbalances. It performs scaffold splits within the same category to preserve label distribution similarity to the original dataset as much as possible. Additionally, to ensure the reliability of experimental conclusions, our dataset partitioning is not fixed and varies with different random seeds. In our experiments, we ensure that data partitions remain consistent under the same random seed.

\begin{figure}[tb]
    \centering
    \includegraphics[width=0.48\textwidth]{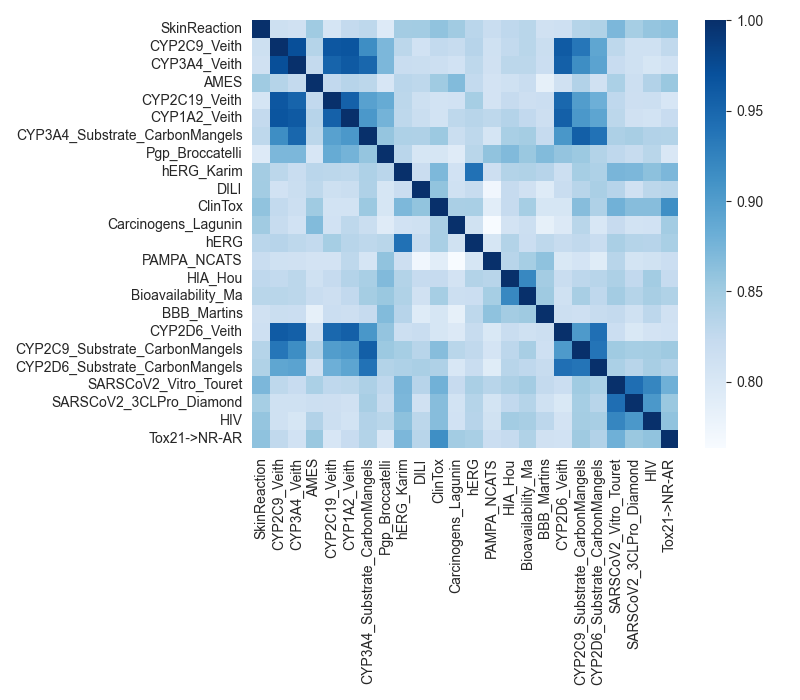}
    \caption{Similarity of Task Description Embeddings}
    \label{fig:sim}
\end{figure}

\subsubsection{Feature Preparation}

The model's input encompass information about the molecular graph and the task identifier. 

For molecular graph information, we need to extract graph from sequence representation, as the majority of datasets represent molecules in the form of SMILES expressions \cite{weininger1988smiles, huang2021therapeutics}. In this paper, we adopt the standard method proposed by the Open Graph Benchmark (OGB) \cite{hu2021ogblsc} to extract graphs from the SMILES and embed molecules into a specified dimension. 

For task identifier, the embedding process of task is relatively more complex. If we simply enumerate all tasks and generate the task embeddings using the one-hot method, like method in \cite{liu2021multi}, it becomes challenging to capture the semantic distinctions and relationships among tasks. In our model, we employ natural language embeddings of task descriptions to guide expert routing. This assists the model in selecting more semantic similar experts. Furthermore, as natural language is human-readable, our model can be easily extended to other datasets.
In this experiment, we utilized the text-embedding-ada-002 model by OpenAI \cite{ryan2022new} to embed task descriptions. The similarity of embeddings across our 24 datasets is depicted in Fig.~\ref{fig:sim}.

\begin{table*}[tb]
    \caption{Detailed Results of 35 tasks Training Experiments (AUCROC $\uparrow$)}
    \centering
    \renewcommand{\arraystretch}{1.2}
    \begin{tabular}{l@{}c|c@{\hspace{-1mm}}c@{\hspace{-2mm}}c@{\hspace{1mm}}c@{\hspace{-2mm}}c@{}c@{\hspace{3mm}}c@{}c}
    \hline
     \multirow{2}{*}{Method}&Dataset & HIV & SARSCoV2\_VT$^{\mathrm{*}}$ & Tox21 & ClinTox & SARSCoV2\_3CD$^{\mathrm{*}}$ & HIA\_Hou & PAMPA$^{\mathrm{*}}$ & CYP2D6\_Veith  \\ [-3pt]
     &Ratio(Total) & 0.034(26320) & 0.057(949) & 0.064(4180) & 0.077(945) & 0.085(563) & 0.149(369) & 0.152(1301) & 0.193(8403)  \\ \hline
     \multirow{6}{*}{Vanilla}&CNN & 0.72±0.034 & 0.525±0.07 & 0.669±0.06 & \underline{0.822±0.053}& 0.645±0.099 & 0.725±0.125 & 0.621±0.061 & 0.795±0.012  \\ 
     &GIN & 0.704±0.045 & 0.554±0.041 & 0.728±0.054 & 0.816±0.055 & 0.719±0.081 & 0.93±0.046 & 0.601±0.042 & 0.81±0.026  \\ 
     &GCN & 0.723±0.08 & 0.501±0.082 & 0.711±0.052 & \textbf{0.842±0.068} & 0.721±0.09 & 0.879±0.051 & 0.615±0.034 & 0.83±0.049  \\ 
     &NeuralFP & 0.751±0.033 & 0.536±0.091 & 0.723±0.051 & 0.819±0.051 & 0.709±0.105 & 0.909±0.047 & 0.631±0.054 & 0.826±0.021  \\
     &Chemprop & 0.748±0.026 & 0.563±0.071 & 0.754±0.047 & \underline{0.836±0.027}& 0.715±0.083 & 0.829±0.049 & 0.659±0.067 & \underline{0.856±0.013}\\
     & AttentiveFP & 0.72±0.052 & 0.534±0.092 & 0.735±0.056 & 0.605±0.121 & \underline{0.732±0.104}& 0.891±0.046 & 0.667±0.046 &0.839±0.017  \\ \hline
     \multirow{3}{*}{Pretrain}&KCL & 0.75±0.029 & 0.577±0.083 & 0.715±0.043 & 0.594±0.069 & 0.713±0.099 & 0.887±0.044 & 0.657±0.035 & 0.801±0.022  \\ 
     &ContextPred & \underline{0.763±0.034}& \underline{0.581±0.057} & \underline{0.78±0.045} & 0.71±0.075 & 0.72±0.107 & \underline{0.932±0.031}& \underline{0.684±0.048}& \textbf{0.902±0.012}\\ 
     &AttrMasking & \underline{0.768±0.041} & \underline{0.599±0.057} & \underline{0.772±0.041} & 0.684±0.076 & \underline{0.729±0.127}& \underline{0.933±0.025}& \underline{0.668±0.047}& \underline{0.897±0.008}\\ \hline 
     ~&GNN-MoCE(Ours) & \textbf{0.779±0.036} & \textbf{0.645±0.09}1 & \textbf{0.808±0.029} & 0.807±0.08 & \textbf{0.765±0.056}& \textbf{0.953±0.034}& \textbf{0.701±0.048}& 0.855±0.017  \\ \hline \hline

     \multirow{2}{*}{Method}&Dataset & CYP2C9\_SC$^{\mathrm{*}}$ & Carcinogens$^{\mathrm{*}}$ & Bio\_Ma$^{\mathrm{*}}$ & BBB\_Martins & CYP2D6\_SC$^{\mathrm{*}}$ & hERG & SkinReaction & CYP2C9\_Veith  \\ [-3pt]
     &Ratio(Total) & 0.203(428) & 0.229(179) & 0.247(409) & 0.248(1299) & 0.293(426) & 0.305(419) & 0.32(231) & 0.336(7738)  \\ \hline
     \multirow{6}{*}{Vanilla}&CNN & \underline{0.617±0.042}& 0.832±0.099 & 0.653±0.075 & 0.835±0.047 & 0.659±0.054 & 0.695±0.056 & 0.146±0.157 & 0.849±0.014  \\ 
     &GIN & 0.544±0.072 & 0.797±0.099 & 0.598±0.07 & 0.818±0.051 & 0.669±0.066 & 0.705±0.054 & 0.177±0.142 & 0.821±0.035  \\ 
     &GCN & 0.589±0.015 & 0.772±0.043 & 0.588±0.044 & 0.852±0.105 & 0.684±0.049 & 0.711±0.072 & 0.243±0.051 & 0.861±0.143  \\ 
     &NeuralFP & 0.565±0.036 & 0.838±0.081 & 0.607±0.063 & 0.838±0.053 & 0.701±0.052 & 0.718±0.068 & 0.193±0.12 & 0.862±0.016  \\
     &Chemprop & 0.592±0.073 & 0.837±0.091 & \underline{0.659±0.066} & \underline{0.857±0.044} & \underline{0.732±0.056} & 0.725±0.051 & 0.109±0.139 & 0.882±0.012  \\ 
     &AttentiveFP & 0.586±0.064 & \underline{0.853±0.072} & \underline{0.655±0.089} & 0.824±0.054 & 0.726±0.055 & \underline{0.796±0.066} & \underline{0.392±0.289} & 0.866±0.015  \\ \hline
     \multirow{3}{*}{Pretrain}&KCL & \underline{0.614±0.074}& 0.821±0.066 & 0.645±0.076 & 0.833±0.069 & 0.725±0.056 & 0.78±0.055 & \textbf{0.52±0.146} & 0.852±0.017  \\ 
     &ContextPred & 0.547±0.075 & 0.834±0.073 & 0.62±0.094 & \underline{0.871±0.043} & \underline{0.731±0.064} & 0.764±0.062 & 0.347±0.185 & \textbf{0.927±0.007}  \\ 
     &AttrMasking & 0.608±0.025 & \underline{0.857±0.071} & 0.625±0.101 & 0.849±0.037 & 0.724±0.056 & \underline{0.792±0.045} & 0.346±0.146 & \underline{0.921±0.011}  \\ \hline
     ~ &GNN-MoCE(Ours) & \textbf{0.644±0.065}& \textbf{0.935±0.028}& \textbf{0.731±0.068} & \textbf{0.902±0.047} & \textbf{0.819±0.042} & \textbf{0.835±0.049} & \underline{0.51±0.076} & \underline{0.902±0.011}  \\ \hline \hline

     \multirow{2}{*}{Method}&Dataset & CYP3A4\_Veith & AMES & CYP2C19\_Veith & CYP1A2\_Veith & CYP3A4\_SC$^{\mathrm{*}}$ & Pgp\_Bro$^{\mathrm{*}}$ & hERG\_Karim & DILI  \\ [-3pt]
     &Ratio(Total) & 0.408(7889) & 0.44(4553) & 0.447(8105) & 0.468(8050) & 0.479(428) & 0.489(779) & 0.492(8604) & 0.497(304)  \\ \hline
     \multirow{6}{*}{Vanilla}&CNN & 0.844±0.021 & 0.739±0.034 & 0.846±0.014 & 0.887±0.008 & 0.571±0.088 & 0.892±0.027 & 0.772±0.015 & 0.791±0.057  \\ 
     &GIN & 0.833±0.023 & 0.752±0.024 & 0.832±0.033 & 0.865±0.027 & 0.526±0.078 & 0.849±0.042 & 0.761±0.024 & 0.77±0.079  \\ 
     &GCN & 0.862±0.02 & 0.79±0.015 & 0.868±0.024 & 0.9±0.008 & 0.585±0.012 & 0.893±0.083 & 0.8±0.021 & 0.8±0.015  \\ 
     &NeuralFP & 0.867±0.019 & 0.796±0.026 & 0.868±0.017 & 0.894±0.014 & 0.548±0.053 & 0.843±0.041 & 0.798±0.016 & 0.823±0.041  \\ 
     &Chemprop & 0.883±0.018 & \underline{0.821±0.017} & 0.884±0.009 & 0.908±0.012 & 0.529±0.068 & 0.759±0.043 & \underline{0.822±0.018} & 0.824±0.065  \\ 
     &AttentiveFP & 0.868±0.022 & 0.803±0.037 & 0.871±0.016 & 0.906±0.013 & \underline{0.607±0.08} & \underline{0.916±0.034} & 0.792±0.017 & \underline{0.828±0.059}  \\ \hline
     \multirow{3}{*}{Pretrain}&KCL & 0.841±0.026 & 0.745±0.03 & 0.843±0.011 & 0.873±0.017 & \underline{0.596±0.077} & \textbf{0.92±0.041} & 0.786±0.021 & 0.791±0.049  \\ 
     &ContextPred & \underline{0.918±0.012} & 0.799±0.02 & \textbf{0.923±0.008} & \textbf{0.939±0.008} & 0.564±0.077 & 0.893±0.034 & 0.816±0.019 & \underline{0.829±0.042}  \\ 
     &AttrMasking & \textbf{0.919±0.012} & \underline{0.806±0.022} & \underline{0.92±0.008} & \underline{0.936±0.008} & 0.552±0.081 & \underline{0.916±0.018} & \underline{0.819±0.012} & \textbf{0.839±0.023}  \\ \hline
     ~ &GNN-MoCE(Ours) & \underline{0.898±0.013} & \textbf{0.834±0.026} & \underline{0.886±0.012} & \underline{0.912±0.009} & \textbf{0.737±0.059} & 0.881±0.041 & \textbf{0.852±0.027} & 0.821±0.062  \\ \hline \\ [-2mm]
     \multicolumn{10}{l}{$^{\mathrm{*}}$ The following are the full names of the dataset abbreviations: ~CYP2C9\_SC (CYP2C9\_Substrate\_CarbonMangels), ~Bio\_Ma ~(Bioavailability\_Ma),} \\
     \multicolumn{10}{l}{SARSCoV2\_VT (SARSCoV2\_Vitro\_Touret), CYP2D6\_SC (CYP2D6\_Substrate\_CarbonMangels), SARSCoV2\_3CD (SARSCoV2\_3CLPro\_Diamond),} \\
     \multicolumn{10}{l}{CYP3A4\_SC (CYP3A4\_Substrate\_CarbonMangels), Carcinogens (Carcinogens\_Lagunin), Pgp\_Bro (Pgp\_Broccatelli), PAMPA (PAMPA\_NCATS)}
    \end{tabular}
    
    \vspace{-0.3cm}
    \renewcommand{\arraystretch}{1.1}
    \label{table:MTL-Exp}
\end{table*}
\begin{table}[tb]
    \centering
    \caption{Average Results of 35 Tasks Training Experiments}
    \begin{tabular}{cc|c}
    \hline
        Method & Model & Mean AUCROC $\uparrow$  \\ \hline
        \multirow{6}{*}{Vanilla} & CNN & 0.715±0.055  \\ 
        ~ & GIN & 0.716±0.054  \\ 
        ~ & GCN & 0.734±0.051  \\ 
        ~ & NeuralFP & 0.736±0.049  \\ 
        ~ & Chemprop & 0.741±0.049  \\
        ~ & AttentiveFP & 0.75±0.063  \\ 
        \hline
        \multirow{3}{*}{Pretrain} & KCL & 0.745±0.052  \\ 
        ~ & ContextPred & \underline{0.766±0.051}  \\ 
        ~ & AttrMasking & \underline{0.77±0.046}  \\ \hline
        ~ & GNN-MoCE(Ours)& \textbf{0.809±0.043}  \\ 
        \hline
    \end{tabular}
    \label{table:MTL-Exp-summary}
    \vspace{-2.5mm}
\end{table}

\subsubsection{Baselines}
We compared our proposed method against 9 common MPP models. Based on training methods, we categorize them into two main groups: Vanilla and Pretrain. Vanilla methods follow the classical model training approach, which involves training on a single dataset. In contrast, Pretrain methods require additional information sources and often leverage large-scale tasks to acquire a foundational model. This model is then fine-tuned on the train and validation sets of the target task and subsequently tested on the test set.

In this experiment, we employed 6 types of Vanilla methods. Among them, CNN and NeuralFP are sequence-based methods, while the remaining 4 are graph-based methods. The details are described as follows:

\begin{itemize}
    \item \textbf{CNN} \cite{huang2020deeppurpose}, a basic convolutional neural network model based on SMILES sequences, improved for MPP. This is not the graph-based method.
    \item \textbf{GCN} \cite{kipf2016semi}, a fundamental GNN structure, which achieves message passing through a simple aggregation function.
    \item \textbf{GIN} \cite{xu2018powerful}, a GNN structure that introduces graph isomorphism testing capability.
    \item \textbf{NeuralFP} \cite{9412489}, a neural network model using autoencoders for each layer, analyzing autoencoder error distributions during the reconstruction of the training dataset.
    \item \textbf{Chemprop} \cite{yang2019analyzing}, a classic deep GNN structure designed for molecular embeddings, augmented with various features that can be rapidly computed in the molecular domain.
    \item \textbf{AttentiveFP} \cite{xiong2019pushing}, a graph neural network structure utilizing graph attention mechanisms learned from relevant drug discovery datasets.
\end{itemize}

For Pretrain MPP methods, we primarily considered the following three pre-training approaches. They integrate various external information into their models.

\begin{itemize}
    \item \textbf{KCL} \cite{fang2022molecular}, a novel knowledge-enhanced contrastive learning (KCL) framework for molecular representation learning, based on a Chemical Element Knowledge Graph and contrastive learning. This introduces the external chemical knowledge into MPP.
    \item \textbf{ContextPred} \cite{hu2019strategies}, an unsupervised pre-training graph model based on context prediction. The model undergoes pre-training on a large-scale dataset to acquire molecular context information. This introduces supplementary molecular structures into MPP.
    \item \textbf{AttrMasking} \cite{hu2019strategies}, an unsupervised pre-training graph model based on masking node attributes. The model undergoes pre-training on a large-scale dataset to enhance node representations. This also introduces supplementary molecular structures into MPP.
\end{itemize}

In contrast to the aforementioned baselines, our model is trained using information across various tasks. Without utilizing additional information sources, we integrate the training and validation sets of the existing 24 datasets. The model is jointly trained on the integrated development set and subsequently tested on the respective test sets.

\subsubsection{Metrics}
All evaluations were conducted based on AUCROC (Area Under Receiver Operating Characteristic Curve) values. A higher value of this metric indicates better performance.

\subsubsection{Training details}
For the optimization strategy, we employed AdamW \cite{loshchilov2017decoupled} with a batch size of 512, setting both the initial learning rate and weight decay to 0.01. To dynamically adjust the learning rate, we implemented a Cosine Annealing Schedule. The architectural composition of our model featured 60 experts in each processing layer, and each Encoder incorporated 6 Graph Neural Networks (GNNs), each characterized by an embedding dimension of 300. The decision group size $k_s$ is set to 4, and task visiable experts number $k_t$ is set to 12. 

In the comparative study, we subjected the baseline models to experimentation using their default parameters as outlined in their respective publications. All experiments were conducted on an 80GB Nvidia A800 GPU accelerator.

\subsection{Main Experimental Results}

\subsubsection{Model Performance} \label{part:full_exp}

We evaluated the performance of the GNN-MoCE model on 35 tasks across 24 datasets and compared it with common MPP models that were separately trained or fine-tuned on each dataset. The detailed experimental results are presented in TABLE~\ref{table:MTL-Exp}, and the average performance of each model across all datasets are presented in TABLE~\ref{table:MTL-Exp-summary}. The numbers in the table represent the mean and standard deviation of the AUC-ROC values obtained from 10 runs with different random seeds. Higher values indicate better performance. The models are grouped based on whether they are pretrained. In each group, they are ranked by their average performance. The best model's result is highlighted in bold, and the top three models are indicated with an underscore.

We found that our model outperformed other baseline models in average performance across all datasets and achieved the best performance in several tasks. Notably, the datasets in TABLE~\ref{table:MTL-Exp} are sorted in ascending order according to the positive class ratio (the number of positive instances divided by the total number of instances). The results show that the tasks with substantial gains relative to the baseline models are mostly the datasets with limited data or extremely imbalanced class distributions.

\subsubsection{Ablation Study}
To evaluate the effectiveness of our architecture, we systematically removed the mechanisms we proposed step by step. Experiments was conducted on the ten datasets, and their average results were presented in TABLE~\ref{tab-ablation}. 
In the table, GNN-MoCE represents the complete model, incorporating mechanisms of Expert-Specific Projection and Expert-Specific Loss. As can be seen from the table, compared to the GNN that does not use our method at all, our GNN-MoCE brings a total AUCROC gain of 0.107.
Specifically, the exclusion of the Expert-Specific Loss incurs a performance loss in AUCROC of 0.041. This operation implies that we no longer consider the dominant decision problem in the dynamic decision group, and the expert group may degenerate into a single expert. Next, we remove the Expert-Specific Projection and find that AUCROC decreases from 0.801 to 0.769. Of this change, 0.019 is contributed by the Attention Cosine Loss in Expert-Specific Projection. This indicates that our approach, which involves assigning each expert a unique perspective for observing samples and ensuring these perspectives are as distinct as possible, has proven to be effective.

\begin{table}[tb]
% \vspace{-3mm}
    \caption{Ablation Study Average Results}
    \centering
\begin{tabular}{lc}
\hline
\textbf{Model} & \textbf{AUCROC $\uparrow$}\\ \hline
GNN-MoCE & 0.842 
\\
GNN-MoCE w/o ES Loss& 0.801
\\
GNN-MoCE w/o (ES Loss \& $\mathcal L_{att}$ )& 0.782 
\\
GNN-MoCE w/o (ES Loss \& ES Projection )& 0.769
\\
GNN & 0.735  \\ \hline
\\ [-2mm]
\multicolumn{2}{l}{$^{\mathrm{1}}$ Evaluated on SkinReaction,~ hERG\_Karim, ~CYP1A2\_Veith} \\
\multicolumn{2}{l}{AMES, CYP2C9\_Veith, CYP3A4\_Veith, Pgp\_Broccatelli, DILI} \\
\multicolumn{2}{l}{CYP2C19\_Veith and CYP3A4\_Substrate\_CarbonMangels} \\
\multicolumn{2}{l}{$^{\mathrm{2}}$ ES denotes Expert-Specific} \\
\multicolumn{2}{l}{$^{\mathrm{3}}$ ES Projection = SAG + $\mathcal L_{att}$}
\end{tabular}
    \label{tab-ablation}
    \vspace{-3mm}
\end{table}
% The results reveal that the removal of the MoCE module results in a 12.70\% decrease in the model's average performance of learning across the selected 10 datasets. 

% Specifically, the exclusion of the Expert-Specific Loss incurs an approximately 4.87\% performance loss, highlighting the significance of enabling fair training for experts in contributing to overall performance improvement. Within the Expert-Specific Projection, we eliminated components aimed at enhancing the distinctiveness of expert perspectives, including the Attention Cosine Loss and the SAG module facilitating individualized expert viewpoints. This removal process collectively contributes to an approximately 4\% performance loss. Furthermore, the measure of enhancing the distinctiveness of expert perspectives demonstrates performance improvement, indicating that diversifying the perspectives through which experts perceive samples is advantageous for collaborative expert learning.

\subsection{Exploratory Experimental Results}

Due to the nature of our task embeddings being derived from natural language, it is relatively simple to extend our model to new tasks. All that is required is to modify the natural language descriptions of tasks and prepare the corresponding datasets. To validate the scalability of this task embedding approach, we retained the default training settings while excluding certain tasks during training. For these tasks, we tested whether the model could achieve reasonably good performance without any additional training, solely relying on the natural language embeddings of the new tasks.
We masked four datasets from the training set for this purpose. Among them, CYP2C19\_Veith and CYP3A4\_Veith are the most similar to the other 31 datasets, while PAMPA\_NCATS and HIA\_Hou are the least similar.

The experimental results are presented in TABLE~\ref{table:mased_task}. In the table, column Masked denotes the model's performance when trained with only 31 tasks and directly tested on these four datasets. Compared to Unmasked, which means training and testing the model with all 35 tasks together, the performance only drops by about 20\%. When compared to the pretrained model without finetuning, which is developed using ContextPred, our model exhibits a performance gain ranging from 7\% to 46\%. This suggests that our model possesses some ability to do OOD tasks.
\begin{table}[bt]
    \centering
    % \vspace{-6mm}
    \caption{Results of masked tasks (AUCROC $\uparrow$)}
    \begin{tabular}{c|ccc}
    \hline
        \textbf{Dataset} & \textbf{Pretrain(ContextPred)} & \textbf{Masked} & \textbf{Unmasked} \\ \hline
        CYP3A4\_Veith & 0.506  & 0.702 & 0.898  \\ 
        CYP2C19\_Veith & 0.492  & 0.743 & 0.886  \\ 
        PAMPA\_NCATS & 0.498  & 0.548 & 0.701  \\ 
        HIA\_Hou & 0.486  & 0.755 & 0.953  \\ \hline
    \end{tabular}
    \label{table:mased_task}
\end{table}
\section{Conclusion}
In this paper, we propose the GNN-MoCE architecture to leverage the collaborative effects among multiple experts corresponding to various MPP tasks. The proposed method for enhancing expert collaboration is woven throughout the entire lifecycle of the dynamic decision group. For the generation of candidate experts, we propose the Expert-Specific Projection. By employing the unique SAG readout parameters for each expert, we strive to obtain distinct graph representations, thereby obtaining as diverse experts as possible. For the formation of a dynamic decision group, the embeddings of task natural language descriptions are utilized in routing samples, thereby ensuring the experts involved in the collaboration are proficient in similar tasks. 
As for the collaboration, the Expert-Specific Loss is introduced to emphasize the fairness of training among the experts within the dynamic decision group. This collaborative decision-making within the group can be seen as a manifestation of swarm intelligence.
Additionally, benefiting from the human-editability of natural language task descriptions, our model exhibits the potential to execute new tasks without the need for additional training.

Considering future research, the encoder in the architecture can be replaced with other pretrained GNN models, requiring only engineering adjustments. Introducing high-quality pre-trained models into the encoder may further improve generalization in OOD tasks. Besides, the proposed MoCE method can be further generalized to other models with MoE structures, allowing MoE-structured models to leverage the collaborative effects within the expert group on top of their original sparse routing structure.

\section*{Acknowledgement}

This work is supported by the National Key Research and Development Program of China under Grant No. 2022YFC3303301, the National Natural Science Foundation of China under Grant No. 72271151, and the Blue Information Entrepreneurship and Innovation Fund of Shanghai University of Finance and Economics.

\bibliographystyle{ieeetr}
\bibliography{ref.bib}

\end{document}